\documentclass[sn-mathphys-num]{sn-jnl}

\usepackage{graphicx}%
\usepackage{multirow}%
\usepackage{amsmath,amssymb,amsfonts}%
\usepackage{amsthm}%
\usepackage{mathrsfs}%
\usepackage[title]{appendix}%
\usepackage{xcolor}%
\usepackage{textcomp}%
\usepackage{manyfoot}%
\usepackage{booktabs}%
\usepackage{algorithm}%
\usepackage{algorithmicx}%
\usepackage{algpseudocode}%
\usepackage{listings}%

\theoremstyle{thmstyleone}%
\theoremstyle{thmstyletwo}%

\theoremstyle{thmstylethree}%
\newtheorem{definition}{Definition}%

\begin{document}

\title[Article Title]{A Network-driven Framework for Public Event Forecasting via Dynamic Interaction Network Evolution}


\author[1]{\fnm{Jie} \sur{Wei}}\email{weijie@stu.cdut.edu.cn}

\author[3]{\fnm{Yue} \sur{Liu}}\email{lyue501@163.com}

\author[1]{\fnm{Xiaochuan} \sur{Tang}}\email{tangchuan@uestc.edu.cn}

\author[1]{\fnm{Biao} \sur{Cai}}\email{caibiao@cdut.edu.cn}

\author[2]{\fnm{Xiangtao} \sur{Li}}\email{lixt314@jlu.edu.cn}

\author*[1]{\fnm{Yanmei} \sur{Hu}}\email{huyanmei@cdut.edu.cn}

\affil*[1]{\orgdiv{Chengdu University of Technology}, \city{Chengdu}, \postcode{610000}, \country{China}}

\affil[2]{\orgdiv{Jilin University}, \city{Changchun}, \postcode{130000}, \country{China}}

\affil[3]{\orgdiv{Chengdu Neusoft University}, \city{Chengdu}, \postcode{610000}, \country{China}} 

\abstract{Effective public event forecasting is essential for intelligent service systems, enabling proactive risk management, adaptive resource allocation, and timely decision-making. In many real-world scenarios, the evolution of public events is driven by dynamic interactions among participants. Motivated by this observation, this paper proposes auto-ibDLM, a network-driven deep learning framework that represents events as dynamic interaction networks and predicts public event evolution through participant growth forecasting. The proposed framework adopts a hybrid representation learning strategy that first represents network evolution using network science-informed structural metrics and subsequently transforms the resulting structural feature vectors into compact and robust latent representations through an auto-learning layer. A GRU-based temporal forecasting module is then employed to capture temporal dependencies and predict future participant growth. Extensive experiments on 13 real-world public event datasets and two publicly available dynamic network datasets demonstrate that auto-ibDLM consistently outperforms representative state-of-the-art methods in both forecasting accuracy and generalization capability, achieving over 97\% accuracy in public event forecasting. Comprehensive experimental analyses further validate the effectiveness of the proposed hybrid representation learning strategy and demonstrate its representation-level interpretability. These results indicate that auto-ibDLM provides an effective and practical solution for intelligent public event forecasting.}


\maketitle

\section{Introduction}\label{sec: introduction}
Public events often involve large numbers of participants within confined spaces and limited time windows, and their rapid escalation can exert substantial pressure on urban infrastructure and public services, potentially threatening public safety. A notable example is the crowd surge during Halloween celebrations in Itaewon, Seoul, which resulted in 159 fatalities and 196 injuries. Such incidents underscore the critical need for timely forecasting of public event evolution to enable proactive decision-making, risk mitigation, and coordinated resource allocation before events escalate into public safety crises.

Significant progress has been made in public event management, including demand-aware resource allocation and distribution~\cite{fei2022demand}, post-disaster assessment and situational awareness~\cite{eckhardt2022using,yigitcanlar2022detecting}, and public opinion monitoring and management~\cite{pelen2022vector}. Despite these advances, existing studies primarily focus on responding to or managing ongoing events rather than anticipating how public events evolve before risks materialize. Consequently, the ability to forecast public event evolution remains limited~\cite{zhao2021event}. This limitation hinders proactive risk management, early warning, and timely intervention before public events escalate into large-scale safety incidents.

For many public events, event evolution is fundamentally driven by participant dynamics, and adverse outcomes are likely to occur when participation intensity exceeds safe capacity limits within a short period~\cite{liang2024unraveling}. For example, consider a Mid-Autumn Festival activity planned for a venue with a safe capacity of 20,000 participants. If forecasts indicate that future participant growth will exceed the venue's safe capacity, the event can be identified as high risk, enabling event organizers and emergency management authorities to deploy proactive measures such as access control, crowd diversion, or resource reinforcement. Since participant growth directly reflects the evolution of many public events, forecasting future participant growth provides an effective means of anticipating public event evolution before critical capacity thresholds are exceeded.

In the digital era, potential participants interact extensively through online platforms by sharing, discussing, and commenting on event-related content. These interactions not only reflect public attention but also influence subsequent participation behaviors, thereby continuously driving the evolution of public events. This perspective naturally motivates representing an event as a dynamic interaction network, where nodes represent participants and edges represent their interactions. Consequently, the evolution of the participant interaction network naturally reflects the evolution of the underlying event, making public event forecasting a task of forecasting the evolution of dynamic interaction networks. This network-centric formulation provides the foundation for forecasting public event evolution by modeling the evolution of dynamic interaction networks.

Building on this network-centric formulation, we propose a network-driven deep learning framework, termed auto-ibDLM (interaction behavior-driven deep learning model), for public event forecasting. Auto-ibDLM represents an event as a dynamic interaction network constructed from participant interactions and formulates public event forecasting as the task of forecasting future participant growth, represented as node increments in the evolving interaction network. Specifically, each network snapshot is represented by a structural feature vector constructed from multi-level network structural metrics, including network-level properties, statistical summaries of node-level metrics, and community evolution behaviors. Given the complex and non-stationary relationship between these handcrafted structural feature vectors and future participant growth, an auto-learning layer is introduced to transform them into compact and robust latent representations. By combining network science-informed structural feature vectors with automatic representation learning, the proposed hybrid representation learning strategy effectively bridges network science knowledge and data-driven representation learning. This design is particularly suitable for public event forecasting, where rapidly evolving interaction patterns and limited event-specific historical data make purely deep learning-based feature learning difficult. The learned latent representations are subsequently fed into a gated recurrent unit (GRU)-based temporal model to capture sequential dependencies and forecast future participant growth. The main contributions of this work are summarized as follows:
\begin{itemize}
\item We propose auto-ibDLM, a network-driven deep learning framework that models events as dynamic interaction networks and formulates public event forecasting as forecasting future participant growth, represented by node increments. To the best of our knowledge, auto-ibDLM explicitly formulates public event forecasting as an evolution forecasting problem on dynamic interaction networks, providing a novel network-centric modeling perspective.

\item We propose a hybrid representation learning strategy that represents each network snapshot using structural feature vectors constructed from network science-informed structural metrics and further transforms them into compact and robust latent representations through an auto-learning layer. The handcrafted structural feature representation reduces the dependence on large-scale event-specific training data, while the learned mapping between structural feature vectors and latent representations naturally provides representation-level interpretability.

\item Extensive experiments on real-world public event datasets and publicly available dynamic network datasets demonstrate the effectiveness, generalizability, and representation-level interpretability of the proposed framework, consistently outperforming representative state-of-the-art methods in public event forecasting.
\end{itemize}

\section{Related Work}\label{sec: related work}
In this work, we formulate public event forecasting as the task of forecasting the evolution of dynamic interaction networks. Accordingly, this section reviews related work in two closely related areas: public event forecasting and dynamic network evolution forecasting.

\subsection{Public event forecasting}
The rise of social media has offered a new way to explore and understand public events~\cite{an2021measuring}. Previous studies mainly focused on analyzing information and public response after events occurred. Zhou et al. and Huang et al. examined textual information on social media to understand requests for disaster relief, which helped in the efficient distribution of supplies~\cite{zhou2022victimfinder,huang2022integrated}. Kanth et al. assessed flood severity using social media data~\cite{kanth2022deep}. Eckhardt et al. conducted an
economic analysis of floods in Brazil through Twitter data~\cite{eckhardt2022using}. An et al. and Zhang et al. predicted the impact of Weibo during public events to help authorities anticipate risks and reduce public panic caused by terror events~\cite{an2023prediction,zhang2020does}. Similarly, Yin et al., Chen and Du, and Zhang et al. used social media data to forecast early trends in public opinion and emotional changes during public events~\cite{yin2020research,chen2023predicting,zhang2020emotional}.

However, these researches mainly focus on post-event response, lacking early warning and forecasting of public events. Social media, as a key platform for information sharing, offers vast real-time data and extensive user engagement, potentially revealing signs of upcoming events~\cite{chopra2022predicting}. Bono et al. used textual data from social media to detect floods in real-time~\cite{bono2022learning}. Nicolas et al. analyzed discussions on social media to gain insights into public opinion and investor~\cite{nicolas2024esg}. Using textual data from social media, Bouzidi et al. proposed a CNN-LSTM model for disaster management, covering both early warning and situational awareness~\cite{bouzidi2022enhancing}. Chopra et al. aimed to find the best NLP model for forecasting disasters from social media tweets~\cite{chopra2022predicting}.

Existing studies have largely overlooked a critical determinant of event evolution: participant interactions. For participant-driven events, the behaviors and interactions among individuals continuously shape event’s dynamics and overall development. To capture this process, we model such events as dynamic interaction networks constructed from observable participant behaviors. Building on this representation, we develop a deep learning framework that forecasts the evolution of these networks, providing a novel and effective approach for forecasting public events.

\subsection{Dynamic network evolution forecasting}
Dynamic network evolution can be characterized at three complementary levels: micro, meso, and macro. Micro-level evolution focuses on individual entities, such as changes in node attributes or interaction formation; meso-level evolution concerns the dynamics of substructures, such as community formation, growth, and dissolution; and macro-level evolution describes global properties of the network, such as network-scale growth patterns, including the evolution of node or edge counts over time.

In state-of-the-art studies, network embedding, predominantly learned via graph neural networks (GNNs), serves as the foundation for a wide range of dynamic network prediction tasks. For example, Lu et al. modeled edge formation as a sequence of chronological events and proposed a temporal attention-based point process to capture micro-level dynamics, while further parameterizing a general dynamics equation with network embeddings to characterize macro-level edge growth patterns~\cite{Yuanfu2019MMDNE}. Huang et al. proposed Tower4GR, which learns user and group embeddings through multi-scale contrastive learning to recommend online groups that users are likely to join in the future~\cite{Huang2025}. EvolveGCN evolves the parameters of a graph convolutional network across time using recurrent units such as GRU or LSTM, enabling effective modeling of dynamic graphs for tasks including link prediction, edge classification, and node classification~\cite{pareja2020evolvegcn}. WinGNN eliminates explicit temporal encoders by adopting a meta-learning strategy with random gradient aggregation windows, integrating frame-wise and window-aware gradients to learn robust and parameter-efficient dynamic graph representations for link prediction~\cite{Zhu2023WinGNNDG}. Beyond these, Yin et al. extended implicit differentiation to dynamic graph settings to reduce memory overhead in spiking GNNs and introduced information compensation mechanisms to mitigate propagation loss~\cite{yin2024dynamic}, while Zhang et al. proposed LFH, a hypergraph learning framework that incorporates dynamic latent hypergraph construction into node embedding to improve node classification and link prediction in dynamic networks~\cite{zhang2025learning}. SympleDyG treats the historical interactions of each node as a temporal ego-graph and tokenizes it into a sequential representation tailored for Transformer architectures to predict future interactions~\cite{wu2024feasibility}. Comprehensive surveys further highlight the rapid progress in dynamic network embedding methods~\cite{feng2026comprehensive}.

Despite these advances, the forecasting of network-scale evolution, particularly node growth, remains relatively underexplored. Although many dynamic network embedding methods can, in principle, be extended to forecast network size by incorporating additional prediction layers, they generally incur substantial computational and memory overhead on large-scale dynamic networks. Moreover, the learned graph representations provide limited representation-level interpretability, making it difficult to identify which structural factors drive the forecasting results. These limitations motivate the need for more efficient and explainable approaches tailored to macro-level evolution modeling in dynamic interaction networks.

\section{Methodology}\label{sec:method}
\begin{figure*}
    \centering
\includegraphics[width= \linewidth]{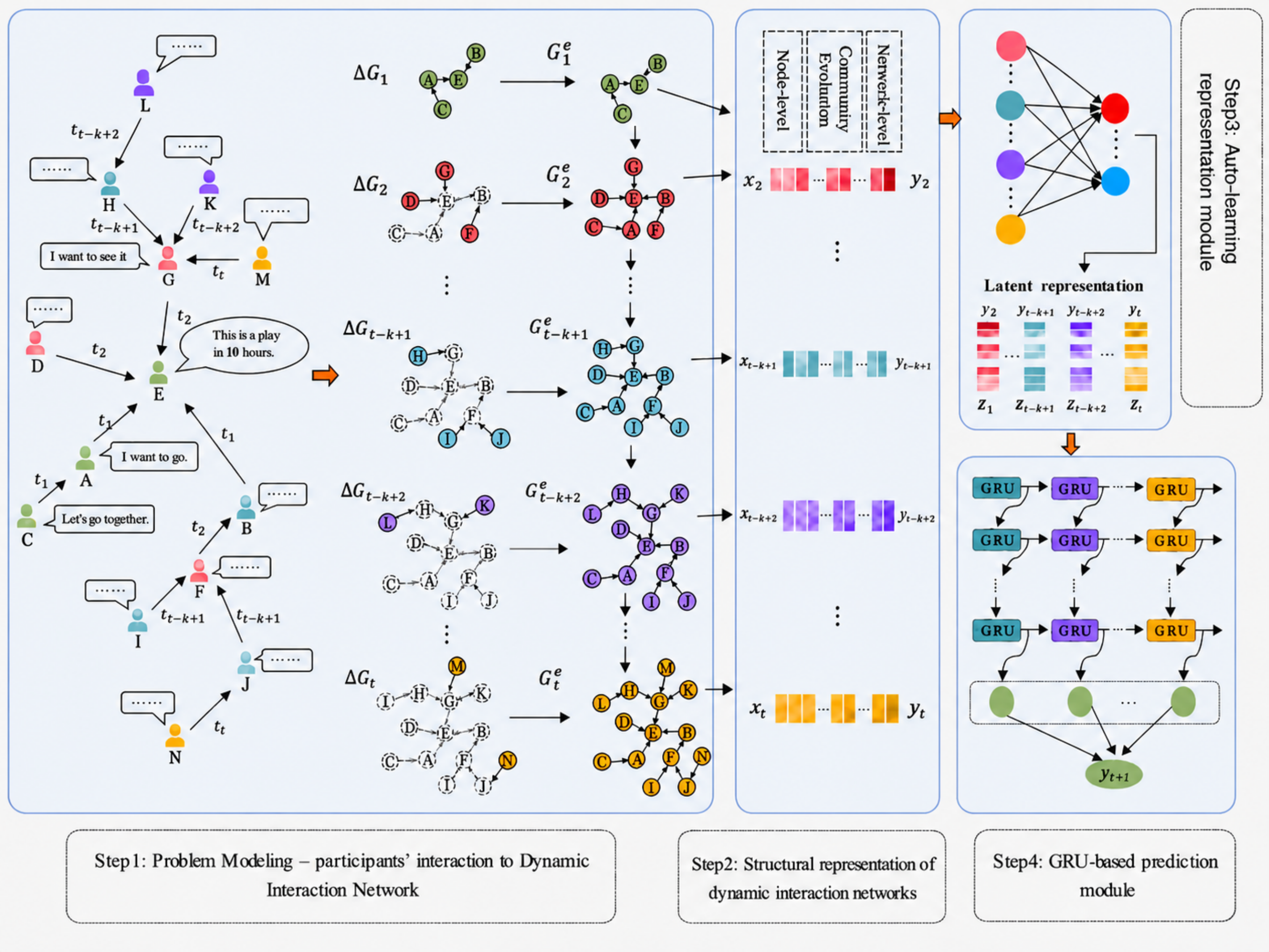}
    \caption{The architecture of auto-ibDLM. The proposed framework represents an event as a dynamic interaction network and formulates public event forecasting as forecasting future participant growth, represented by node increments in the evolving interaction network. The framework consists of four stages: problem modeling, structural representation of the dynamic interaction network, auto-learning representation, and GRU-based temporal forecasting.} 
    \label{fig:model}
\end{figure*}

The overall architecture of auto-ibDLM is illustrated in Fig.~\ref{fig:model}. The proposed framework consists of four stages for network-driven public event forecasting. Given an event, \textbf{Step 1} formulates public event forecasting as the task of forecasting future participant growth by modeling the event as a dynamic interaction network constructed from participants and their interaction behaviors (Section~\ref{subsec:pr}). \textbf{Step 2} represents each network snapshot as a structural feature vector constructed from multi-level network structural metrics (Section~\ref{subsec:cd}). \textbf{Step 3} transforms the structural feature vectors into compact and robust latent representations through an auto-learning layer (Section~\ref{subsec::al}). \textbf{Step 4} employs a GRU-based temporal prediction module to forecast future participant growth represented by node increments (Section~\ref{subsec::gru}). Steps 2 and 3 together constitute the proposed hybrid representation learning strategy, which combines network science-informed structural feature representations with automatic representation learning to generate compact and robust latent representations while maintaining representation-level interpretability.

\subsection{Problem modeling}\label{subsec:pr}
As discussed in the Introduction, the evolution of many events is fundamentally driven by participant dynamics. To capture this evolution from a network-centric perspective, we represent an event as a dynamic interaction network constructed from participants and their interaction behaviors, whose evolution reflects the evolution of the event over time. Based on this formulation, public event forecasting is transformed into forecasting future participant growth represented by node increments in the dynamic interaction network. To formally describe the proposed formulation, we first define an event and its corresponding dynamic interaction network.
\begin{definition}[Event]:
    An event is represented as $e=\{\mathcal{V}^e, \mathcal{E}^e, Attr_{\mathcal{V}}^e, Attr_{\mathcal{E}}^e\}$, where:
    \begin{itemize}
        \item [$\bullet$] $\mathcal{V}^e$ is the set of participants.
        \item [$\bullet$] $\mathcal{E}^e$ is the set of participant interactions.
        \item [$\bullet$] $Attr_{\mathcal{V}}^e$ contains participant attributes (including the timestamp when each participant joins the event).
        \item [$\bullet$] $Attr_{\mathcal{E}}^e$ contains interaction attributes (including the timestamp when each interaction occurs).
    \end{itemize}
    \label{defe}
\end{definition}

\begin{definition} [Dynamic Interaction Network]:
    For an event $e$ observed up to time step $t$, its dynamic interaction network is
    \begin{equation}
       \mathcal{G}^e=(\Delta\mathcal{G}_1^e, \Delta\mathcal{G}_2^e, ..., \Delta\mathcal{G}_{t}^e)
    \label{eq:din} 
    \end{equation}
    where each temporal slice $\Delta\mathcal{G}_i^e=(\Delta\mathcal{V}_i, \Delta\mathcal{E}_i)$ consists of
    \begin{itemize}
    \item [$\bullet$] $\Delta\mathcal{V}_i \subseteq \mathcal{V}^e$: the new participants at time step $i$
    \item [$\bullet$] $\Delta\mathcal{E}_i \subseteq \mathcal{E}^e$: the new participant interactions at time step $i$.
    \end{itemize}
    \label{defd}
\end{definition}

\begin{definition} [Snapshot of Dynamic Interaction Network]:
    The snapshot of the dynamic interaction network at time step $t$ is the accumulation of all participants and interactions up to that time: 
    \begin{equation}
    \mathcal{G}_t^e=\sum_{i=1}^t{\Delta\mathcal{G}_i^e}
                   =\mathcal{G}_{t-1}^e+\Delta\mathcal{G}_t^e
    \label{eq:sp}
    \end{equation}
     with $\mathcal{G}_0^e$ initialized as empty.
\end{definition}

\begin{figure}
    \centering
    \includegraphics[width=0.9\linewidth]{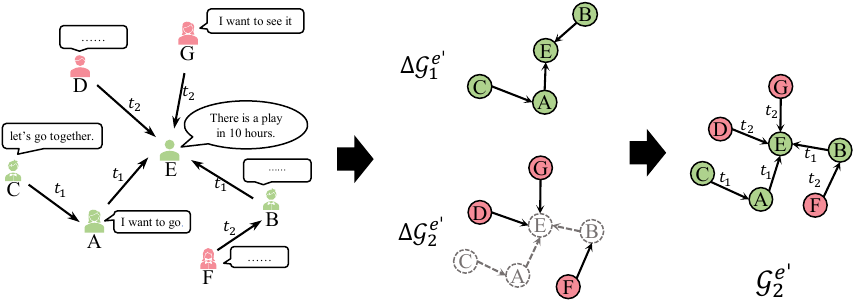}
    \caption{An example of the dynamic interaction network.}
    \label{fig:event-graph}
\end{figure}

Fig.~\ref{fig:event-graph} illustrates an example of the proposed dynamic interaction network. A theater publishes a post on Sina Weibo announcing that a play will begin in ten hours, initiating an event $e'$. The event is observed at an hourly granularity. During the first hour, users A, B, and C participate by replying to, forwarding, and commenting on the post, respectively. During the second hour, users D, F, and G further participate through subsequent interactions. Consequently, the dynamic interaction network observed up to the second hour is
$\mathcal{G}^{e'}=(\Delta\mathcal{G}_1^{e'},\Delta\mathcal{G}_2^{e'})$,
and its corresponding snapshot $\mathcal{G}_2^{e'}$ is illustrated in Fig.~\ref{fig:event-graph}.

Since participant growth directly reflects the evolution of an event, forecasting future participant growth is adopted as the surrogate task for public event forecasting.
\begin{definition} [Network-driven public event forecasting]: Let an event $e$ be observed up to time step $t$, and let $y_{t+1} = |\Delta\mathcal{V}_{t+1}|$ denote the node increment, i.e., the number of new participants that join at time step $t+1$.
\item \textit{Input}: the dynamic interaction network $\mathcal{G}^e$.
\item \textit{Output}: the predicted node increment $\hat{y}_{t+1}$.
\item \textit{Goal}: determine whether the event will evolve into a public event at time step $t+1$ by checking whether $\hat{y}_{t+1}$ exceeds a threshold $\delta$.
\item Formally,
\begin{equation}
    P_e=\mathbf{1}((\hat{y}_{t+1}=\mathcal{F}(\mathcal{G}^e|\boldsymbol{\theta}))>\delta)
    \label{eq:orimodel}
\end{equation}
where $\mathcal{F}$ is a mapping that projects the dynamic interaction network into a predicted node increment.
\end{definition}
In practice, participant growth is mainly influenced by recent interaction dynamics rather than the entire historical evolution. Therefore, we assume a $k$-order Markov property and reformulate Eq.~(\ref{eq:orimodel}) as
\begin{equation}
    P_e=((\hat{y}_{t+1}=\mathcal{F}(\mathcal{G}_{t-k+1}^e,\mathcal{G}_{t-k+2}^e,...,\mathcal{G}_t^e|\boldsymbol{\theta}))>\delta)
    \label{eq:midmodel}
\end{equation}
where only the most recent $k$ network snapshots are used for forecasting.

\subsection{Structural representation of dynamic interaction networks} 
\label{subsec:cd}
To forecast future participant growth, each snapshot of the dynamic interaction network should first be transformed into a structural feature vector that effectively characterizes its evolutionary state. Although graph neural networks (GNNs) have achieved remarkable success in graph representation learning, they often incur substantial computational and memory overhead on large-scale dynamic networks. More importantly, graph-level embeddings learned directly from entire networks generally lack interpretability, making it difficult to understand which structural characteristics are captured and how they contribute to the forecasting results. To address these limitations, we construct network science-informed structural feature vectors from multi-level structural metrics. To characterize network evolution, temporal variations of these structural features between consecutive snapshots are computed and further integrated with community evolution behaviors, yielding a unified structural feature vector for each temporal transition.

\textbf{Network-level structural metrics}.
Given a snapshot $\mathcal{G}_i^e$ of the dynamic interaction network, we characterize its global structural properties using three widely adopted network-level structural metrics: diameter, average shortest path length, and network density. These metrics are formally defined as follows:

\begin{equation}
 diameter(\mathcal{G}_i^e)=max_{(u,v)\in\mathcal{E}_i}{distance(u,v)}
\label{eq:diameter}
\end{equation}

\begin{equation}
    \overline{distance}(\mathcal{G}_i^e)=\sum_{(u,v)\in\mathcal{E}_i}{distance(u,v)}
    \label{eq:distance}
\end{equation}

\begin{equation}
    density(\mathcal{G}_i^e)=\frac{|{E}_i|}{|{V}_i| * (|{V}_i| - 1)}
    \label{eq:density}
\end{equation}
where $distance(u,v)$ denotes the length of the shortest path between nodes $u$ and $v$ in $\mathcal{G}_i^e$. These three metrics are concatenated into a three-dimensional vector, denoted by $\mathbf{x}_i^{g}$, which serves as the network-level component of the structural feature vector for snapshot $\mathcal{G}_i^e$.

\begin{table}[t]
    \centering
     \caption{Node-level structural metrics ($N(u)$ is the set of $u$'s neighbors, and $N^+(u)$ is the combination of $N(u)$ and $u$). Details of these metrics are referred to~\cite{hu2022exhaustive}.}
     \label{tab:node-level}
    \begin{tabular*}{\textwidth}{@{\extracolsep{\fill}} ll}
    \toprule
    Metric & Formula \\ \midrule
    D & $D(u)= \left | N_{u} \right |$  \\
    EXTD & $EXTD(u)= d_{u}+\sum_{v\in N(u)}d_{v}$  \\
    ACCD & $ACCD(u)= d_u+\sum_{v\in N(u)}(d_{v}+\sum_{w\in N(v)}d_{w})$  \\
    NM & $NM(u)= \left | \left \{ (v,w)\in \mathcal{E}\mid v,w\in N^{+}(u) \right \}\right |$  \\
    CE & $CE(u)= \frac{\left | \left \{ \left ( v,w \right ) \in \mathcal{E}\mid v\in N^{+}\left ( u \right ) ,w\in \mathcal{V}-N^{+}\left ( u \right ) \right \}  \right | }{min\left ( vol\left ( N^{+}\left ( u \right ) \right ) \right ),vol\left ( \mathcal{V}-N^{+}\left ( u \right ) \right ) }$  \\
    DE & $DE(u)= \frac{\left | \left \{ (v,w)\in \mathcal{E}\mid v,w\in N^{+}(u) \right \}  \right | }{ \left | \left \{ \left ( v,w \right ) \mid v,w\in N^+\left ( u \right )  \right \}  \right | }$  \\
    LCC & $LCC(u)= \frac{\left | \left \{ \left ( v,w \right ) \in \mathcal{E}\mid v,w\in N\left ( u \right ) \right \}  \right | }{ \left | \left \{ \left ( v,w \right ) \mid v,w\in N\left ( u \right )  \right \}  \right |  }$  \\
    COREDC & $COREDC(u)= \sum_{v\in N(u)} \frac{\left | N(u)\cap N(v) \right | }{\sqrt{\left | N(u)\parallel N(v) \right | } }$  \\
    COREDJ & $COREDJ(u)= \sum_{v\in N(u)} \frac{\left | N(u)\cap N(v) \right | }{\left | N(u)\cup N(v) \right | }$  \\
    COREDP & $COREDP(u)= \frac{\sum_{w}\left ( A_{u,w}-\bar{A}_{u} \right ) \left (A_{v,w}-\bar{A}_{v} \right )}{\sqrt{\sum_{w}\left (A_{u,w}-\bar{A}_{u} \right )^{2} } \sqrt{\sum_{w}\left (A_{v,w}-\bar{A}_{v} \right )^{2} } }$  \\
    COREDPA & $SPA(u)= \sum_{v\in N\left ( u \right ) }d_{u}\times d_{v}$  \\ 
    \botrule
    \end{tabular*}
\end{table}

\textbf{Node-level structural metrics}. We consider 11 widely used node-level structural metrics, as summarized in Table~\ref{tab:node-level}. Since the dynamic interaction network is directed, each metric is computed for both incoming and outgoing connections. Directly incorporating metric values of individual nodes is computationally inefficient and results in feature vectors whose dimensionality varies with the network size. Moreover, node-level metrics often exhibit considerable fluctuations across individual nodes, making the resulting representation sensitive to local variations. Therefore, for each metric, we compute its mean and standard deviation over all nodes, denoted as \emph{Mean} and \emph{Std}, respectively. These aggregated statistics provide a compact yet informative characterization of the node-level structural properties of each snapshot. Consequently, each snapshot is represented by a 44-dimensional node-level structural feature vector (11 metrics $\times$ 2 directions $\times$ 2 statistics), denoted as $\mathbf{x}_i^{l}$.

\textbf{Community evolution behaviors}. Besides structural metrics, community evolution behaviors provide complementary information about the temporal evolution of the dynamic interaction network. In general, seven community evolution behaviors are defined: Birth, Maintain, Dissolve, Merge, Split, Grow, and Shrink~\cite{hu2016local}. Since Birth depends on communities appearing in future snapshots and is therefore unavailable during forecasting, it is excluded from our representation.

To quantify the remaining six behaviors, community detection is first performed on each network snapshot, after which communities in two consecutive snapshots are matched to identify the evolution behavior of each community. Let $\mathcal{C}_{i}$ and $\mathcal{C}_{i+1}$ denote the community sets of $\mathcal{G}_{i}^{e}$ and $\mathcal{G}_{i+1}^{e}$, respectively. For each community $c\in\mathcal{C}_{i}$, we identify its matched community $c'\in\mathcal{C}_{i+1}$ and determine its evolution behavior. The evolution degree is defined as $|c'|-|c|$. For each evolution behavior, we record both the number of communities exhibiting that behavior and the sum of their corresponding evolution degrees, yielding two descriptors per behavior. Consequently, the community evolution between time steps $i$ and $i+1$ is represented by a 12-dimensional vector, denoted as $\mathbf{x}_i^{c}$.

\textbf{Structural feature vector}. The three complementary components described above are integrated to construct the structural feature vector for each temporal transition. Specifically, to characterize the structural evolution of the dynamic interaction network from time step $i$ to time step $i+1$, we compute the temporal variations of the network-level and node-level structural feature vectors, i.e., $\mathbf{x}_{i+1}^{g}-\mathbf{x}_{i}^{g}$ and $\mathbf{x}_{i+1}^{l}-\mathbf{x}_{i}^{l}$, respectively, and concatenate them with the community evolution vector $\mathbf{x}_{i}^{c}$. The resulting structural feature vector is defined as

\begin{equation}
\begin{aligned}
    &\textbf{x}_{i+1}=[\textbf{x}_{i+1}^g-\textbf{x}_{i}^g; \textbf{x}_{i+1}^l-\textbf{x}_{i}^l; \textbf{x}_{i}^c]
    \label{eq:fv}
\end{aligned}
\end{equation}
Accordingly, the evolution of the dynamic interaction network observed up to time step $t$ is represented as the sequential structural feature representation
\begin{equation}
\begin{aligned}
    &\textbf{D}_{\mathcal{G}^e}=((\textbf{x}_{2},y_{2}), (\textbf{x}_{3},y_{3}),...,(\textbf{x}_{t},y_{t}))
    \label{eq:evolution}
\end{aligned}
\end{equation}
where each pair $(\mathbf{x}_{i},y_{i})$ consists of the structural feature vector describing the transition from time step $i-1$ to $i$ and the corresponding node increment.

\subsection{Auto-learning representation module} 
\label{subsec::al}
\textbf{Empirical observations.} After representing the evolution of a dynamic interaction network as a sequence of structural feature vectors, a straightforward solution is to directly feed these vectors into a temporal prediction model such as a recurrent neural network (RNN). However, our empirical analysis on multiple public events from the Sina Weibo platform reveals that different structural features exhibit substantially different temporal relationships with future participant growth. Consequently, their contributions to forecasting are highly heterogeneous.

\begin{figure}
    \centering        \includegraphics[width=\linewidth]{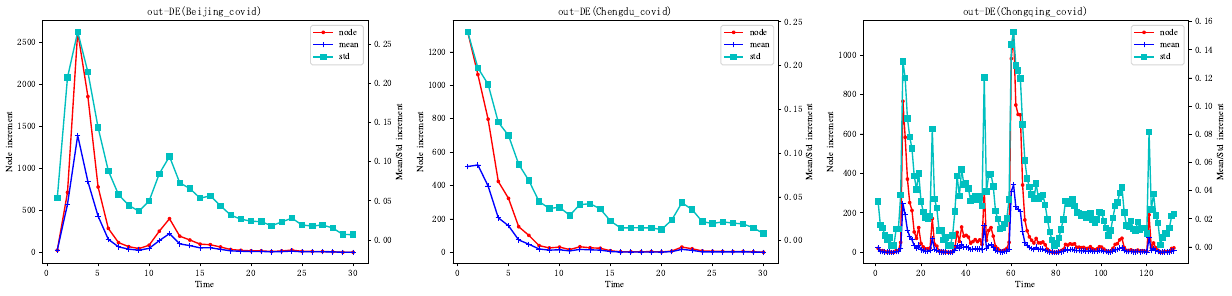} 
    \caption{Temporal variations of the out-DE structural metric (Mean and Std) and the corresponding participant and interaction growth. The horizontal axis represents time steps. The left vertical axis represents participant growth (node increment) and interaction growth (edge increment), while the right vertical axis represents the temporal variations of the Mean and Std of the out-DE structural metric.}
\label{fig:micro_outDE_change}
\end{figure}

\begin{figure}
    \centering       
    \includegraphics[width=\linewidth]{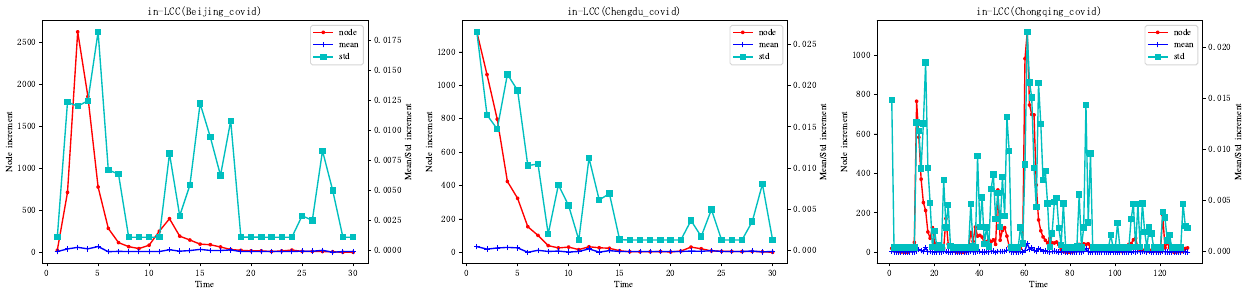}
    \caption{Temporal variations of the in-LCC structural metric (Mean and Std) and the corresponding participant and interaction growth. The horizontal axis represents time steps. The left vertical axis represents participant growth (node increment) and interaction growth (edge increment), while the right vertical axis represents the temporal variations of the Mean and Std of the in-LCC structural metric.}
\label{fig:micro_inLCC_change}
\end{figure}

Figs.~\ref{fig:micro_outDE_change} and \ref{fig:micro_inLCC_change} present two representative examples. The Mean and Std of the ``out-DE" metric generally evolve consistently with the node increment, although local deviations exist. In contrast, the Mean and Std of the ``in-LCC" metric exhibit substantially different temporal trends and only occasionally correlate with the node increment. Similar observations can be found for the remaining structural metrics (see the Supplementary Information\footnote{https://github.com/spidermandark123/Supplementary-Information-for-auto-ibDLM.git\label{fn:supplementaryurl}}). These observations indicate that structural features differ considerably in their predictive relevance and that such relevance cannot be reliably determined through manual feature selection alone.

Manually selecting a subset of seemingly informative features (e.g., retaining only features exhibiting trends similar to ``out-DE'') provides a simple baseline, but suffers from three major limitations. First, partially informative features may be mistakenly discarded because their predictive relevance is only manifested during specific temporal periods. Second, local fluctuations can easily obscure the underlying temporal correlations between structural features and participant growth. Third, feature relevance varies across different public events, making manually designed feature selection difficult to generalize. Therefore, rather than explicitly selecting structural features, we introduce an auto-learning layer that automatically learns a compact and robust latent representation from the structural feature vectors.

Formally, the auto-learning layer is implemented as a fully connected linear transformation:

\begin{equation}
\textbf{z}_{i} = \textbf{W}_{a} \textbf{x}_{i}
\label{eq: latent}
\end{equation}

where $\mathbf{x}_{i}$ denotes the structural feature vector defined in Eq.~\ref{eq:fv} with dimension $d=59$, $\mathbf{z}_{i}$ denotes the learned latent representation with dimension $d'<d$, and $\mathbf{W}_{a}\in\mathbb{R}^{d'\times d}$ is the learnable transformation matrix. To encourage the learned latent representation to preserve structural information that is predictive of future participant growth, we design the following consistency loss:

\begin{equation}
{loss_f} = \frac{1}{d'k(T-k-2)} \sum_{t=k}^{T}\sum_{j=1}^{d'} \sum_{i=t-k+2}^{t-1} \quad \left[ \left( y_{i+1} - z_{i+1, j} \right) - \left( y_{i} - z_{i, j} \right) \right]^2
\label{eq:loss_f}
\end{equation}

where $T$ is the number of observed snapshots and $T-k-2$ denotes the number of effective training samples when the most recent $k$ snapshots are used for forecasting.

The underlying intuition of Eq.~\ref{eq:loss_f} is to preserve temporal consistency between the latent representation and participant growth. Specifically, each latent dimension is encouraged to evolve consistently with the corresponding node increment. If the participant growth increases (or decreases) between two consecutive time steps, the corresponding latent representation is encouraged to exhibit a similar temporal variation. The term $\left(y_{i+1} - z_{i+1, j}\right) - \left( y_{i} - z_{i, j} \right)$ quantifies the inconsistency between these two temporal variations. Minimizing this inconsistency enables the auto-learning layer to automatically emphasize structurally informative features while suppressing feature components whose temporal dynamics are weakly related to future participant growth, thereby producing compact and robust latent representations for subsequent temporal forecasting.

\subsection{GRU-based prediction module}
\label{subsec::gru}
After obtaining the latent representations $\mathbf{z}$ from the auto-learning representation module, we employ a GRU-based temporal forecasting module to predict future participant growth. GRU is adopted because it can capture temporal dependencies with fewer parameters than Long Short-Term Memory (LSTM) networks, which improves computational efficiency while retaining strong sequence modeling capability. Given the most recent $k$ observations, the module takes the latent representations together with the corresponding historical node increments as input and outputs the predicted node increment $\hat{y}_{t+1}$ through a fully connected layer, as illustrated in Step 4 of Fig.~\ref{fig:model}.

A GRU unit consists of an update gate, a reset gate, a candidate hidden state, and a hidden state. Taking the $i$-th GRU unit in the first hidden layer as an example, we define its input as $\mathbf{Z}_{t-k+1+i} =(\mathbf{z}_{t-k+1+i},y_{t-k+1+i})$, where $\mathbf{z}_{t-k+1+i}$ is the latent representation learned by the auto-learning layer and $y_{t-k+1+i}$ is the corresponding historical node increment. The update gate $\mathbf{u}_{i}$ is computed as follows:

\begin{equation}
    \textbf{u}_{i}=sigmoid(\textbf{W}_{u}[\textbf{h}_{i-1},\textbf{Z}_{t-k+1+i}])
    \label{eq:zt}
\end{equation}

The reset gate $\mathbf{r}_{i}$ is calculated as:

\begin{equation}
    \textbf{r}_{i}=sigmoid(\textbf{W}_{r}[\textbf{h}_{i-1},\textbf{Z}_{t-k+1+i}] )
    \label{eq:rt}
\end{equation}

The candidate hidden state $\tilde{\textbf{h}}_{i}$ is calculated as:

\begin{equation}
    \tilde{\textbf{h}}_{i}=tanh(\textbf{W}_{c}[\textbf{r}_{i}\odot \textbf{h}_{i-1},\textbf{Z}_{t-k+1+i}])
    \label{eq:h_ht}
\end{equation}
The hidden state $\textbf{h}_{i}$ is calculated as:
\begin{equation}
    \textbf{h}_{i}=\textbf{u}_{i}\odot \tilde{\textbf{h}}_{i}+(\textbf{1}-\textbf{u}_{i})\odot \textbf{h}_{i-1}
    \label{eq:ht}
\end{equation}

In the above operations, $\mathbf{h}_{i-1}$ denotes the hidden state of the $(i-1)$-th GRU unit, and $\mathbf{W}_{u}$, $\mathbf{W}_{r}$, and $\mathbf{W}_{c}$ are learnable weight matrices for linear transformations. After the input sequence passes through the stacked GRU layers, the hidden states of the $k-1$ GRU units in the final layer are concatenated to form $\mathbf{H}$, which is then mapped to the predicted node increment $\hat{y}_{t+1}$ through a fully connected layer:
\begin{equation}
    \hat{y}_{t+1} = \textbf{W}\textbf{H}+b
    \label{eq:y_t+1}
\end{equation}
where $\mathbf{W}$ and $b$ denote the weight matrix and bias, respectively. 

\subsection{Model training and inference}
\textbf{Training strategy}. After preprocessing the dynamic interaction network into the sequential structural representation defined in Eq.~\ref{eq:evolution}, auto-ibDLM is trained in two stages. In the first stage, the auto-learning representation module is pre-trained by minimizing the consistency loss $loss_f$ (Eq.~\ref{eq:loss_f}), enabling the latent representation to preserve structural information that is predictive of future participant growth. In the second stage, the entire framework, including both the auto-learning representation module and the GRU-based temporal forecasting module, is jointly optimized by minimizing the mean squared error (MSE) between the predicted and ground-truth node increments:

\begin{equation}
     loss_{m} = \frac{1}{T-k-2}\sum_{t=k}^{T-1}(\hat{y}_{t+1}-y_{t+1})^2
\end{equation}

All learnable parameters are optimized jointly through back-propagation.

\textbf{Inference}. Given an event $e$, participant interactions are first collected and transformed into a dynamic interaction network, from which a sequence of structural feature vectors is constructed according to Eq.~\ref{eq:evolution}. The structural feature vectors from the most recent observation window are then fed into the auto-learning representation module to obtain the corresponding latent representations, which are subsequently processed by the GRU-based temporal forecasting module to predict the node increment $\hat{y}_{t+1}$. Finally, $\hat{y}_{t+1}$ is compared with the predefined threshold $\delta$ to determine whether the event is likely to evolve into a public event.

\section{Experiments}
This section systematically evaluates auto-ibDLM from three complementary perspectives. First, we assess its effectiveness in public event forecasting using a real-world Weibo benchmark. Second, we evaluate its generalizability by predicting node increments on two publicly available dynamic network datasets. Finally, we conduct an ablation study to investigate the contributions of the proposed hybrid representation learning strategy and different categories of structural metrics. Unless otherwise specified, all methods were evaluated under identical preprocessing, training, and evaluation settings.

\subsection{Experimental settings}
\subsubsection{Datasets}
To evaluate public event forecasting, we constructed a benchmark dataset consisting of 13 real-world hot topics collected from Sina Weibo. These topics cover major public concerns (e.g., COVID-19), social incidents (e.g., the ``Running Man'' event in Chengdu), and high-attention tourism congestion events. For each topic, we collected participant interactions, including posting, commenting, liking, and reposting, and constructed dynamic interaction networks following Step 1 in Fig.~\ref{fig:model}. The brief information of these 13 topics is shown in Table ~\ref{tab:weibo-datasets}.

\begin{table}
\centering
\small
\caption{Summary of public events in Weibo benchmark dataset}
\begin{tabular*}{\textwidth}{@{\extracolsep{\fill}} c p{5cm} ccc}
\toprule
Events ID & Events Topic & Time interval & \#Node & \#Edge \\
\midrule
E1 & This round of the epidemic has spread to 19 provinces & 4 days & 3659 & 2951 \\
E2 & The whole city of Xuzhou suspends offline teaching & 4 days & 5749 & 3636 \\
E3 & A special case of a patient infecting thirteen patients appeared in Chengdu & 7 days & 5126 & 7026\\
E4 & Advise Beijing citizens not to go out tomorrow unless necessary & 4 days & 9996 & 8149 \\
E5 & Xuzhou city-wide subway suspension & 3 days & 7355 & 5214 \\
E6 & Shanghai epidemic & 5 days & 5288 & 6167 \\
E7 & Do not open the package when you receive the express from this company in Hebei & 3 days & 29157 & 74167 \\
E8 & Chongqing epidemic & 6 days & 11815 & 17574 \\
E9 & Omicron & 3 days & 19969 & 49776\\
E10 & Watching the flag-raising ceremony at Tiananmen Square & 2 days & 2133 & 2666 \\
E11 & Chengdu runner & 5 days & 3315 & 5451 \\
E12 & Traffic congestion in Henan scenic spots & 3 days & 1787 & 1976 \\
E13 & Traffic congestion in Xian scenic areas & 3 days & 3659 & 2951 \\
\botrule
\end{tabular*}
\label{tab:weibo-datasets}
\end{table}

Each dynamic interaction network was partitioned into hourly snapshots. 
Following Eq.~\ref{eq:evolution}, the evolution of each event was represented as a sequence of structural feature vectors and corresponding node increments. A sliding window containing nine consecutive pairs was adopted, where the first eight pairs served as the input sequence and the ninth pair as the prediction target. Since each individual event yielded only a limited number of samples, all topics were combined into a unified benchmark, denoted as \textit{Weibo}, to learn common evolution patterns across events on the same social platform. To prevent temporal information leakage, samples within each topic were chronologically divided into training (the first 90\%) and testing (the remaining 10\%) subsets before merging. The training subsets from all topics were then aggregated to form the global training set, while the corresponding testing subsets constituted the global test set.

To further evaluate the generalizability of auto-ibDLM beyond public event forecasting, we additionally conducted experiments on two publicly available dynamic network datasets from SNAP\footnote{https://snap.stanford.edu/}: \textit{Superuser} (194,085 nodes and 1,443,339 edges; daily snapshots) and \textit{Bitcoin} (5,881 nodes and 35,592 edges; weekly snapshots). Compared with the surge-driven participant growth observed in Weibo, these datasets exhibit relatively smooth network evolution, providing a complementary benchmark for evaluating node increment forecasting under more general dynamic network scenarios.

\subsubsection{Baselines}
Due to the lack of established benchmarks for public event forecasting on dynamic interaction networks, we selected representative methods from three categories for comparison.

\textbf{Sequence models:} Transformer~\cite{vaswani2017attention}, LSTM~\cite{chen2023predicting}, and CNN--LSTM~\cite{bouzidi2022enhancing}, all of which received the same input sequences of structural feature vectors as auto-ibDLM and were optimized using $loss_m$.

\textbf{Dynamic graph representation learning models:} \\ EvolveGCN~\cite{pareja2020evolvegcn} (both H and O variants), WinGNN~\cite{Zhu2023WinGNNDG} and SiGNN~\cite{chen2025signn}. These models were fed edge lists from $k$ historical snapshots, after which a task-specific MLP (EvolveGCN) or a fully connected layer (WinGNN and SiGNN) was trained using $loss_m$.

\textbf{Statistical time-series models:} VAR~\cite{lutkepohl2005new} and ARIMA~\cite{cai2021study}. VAR used the multivariate feature sequence, while ARIMA operated solely on historical node increments.

\subsubsection{Implementation details}
auto-ibDLM, Transformer, LSTM, CNN--LSTM, VAR, and ARIMA were implemented in Python. EvolveGCN,  WinGNN and SiGNN were evaluated using their official implementations. All models were trained for 3000 epochs using Adam (learning rate $10^{-3}$); EvolveGCN, WinGNN and SiGNN adopted early stopping with a patience of 50 epochs. For auto-ibDLM, the auto-learning dimension was set to $d' = 10$. The prediction module contained eight hidden layers, each with seven GRU cells (since $k-1=7$) and ten channels. The auto-learning layer was pretrained for 2000 epochs and jointly fine-tuned for another 3000 epochs. For community detection, we employed LPA~\cite{garza2019community} and LMC--APR~\cite{hu2022exhaustive}. Community evolution was tracked by matching community sets across consecutive snapshots~\cite{hu2016local}. All experiments were run on a machine with an Intel Core i9-9900K CPU, an NVIDIA GTX 1650 GPU, and 32 GB RAM. In addition, we developed a web-based demonstration system for auto-ibDLM. Details are provided in the Supplementary Information\footref{fn:supplementaryurl}.

\subsection{Public Event Forecasting}
\label{subsec:exp-public-event}
\subsubsection{Overall forecasting performance}
\begin{figure}
    \centering  
    \includegraphics[width=\linewidth]{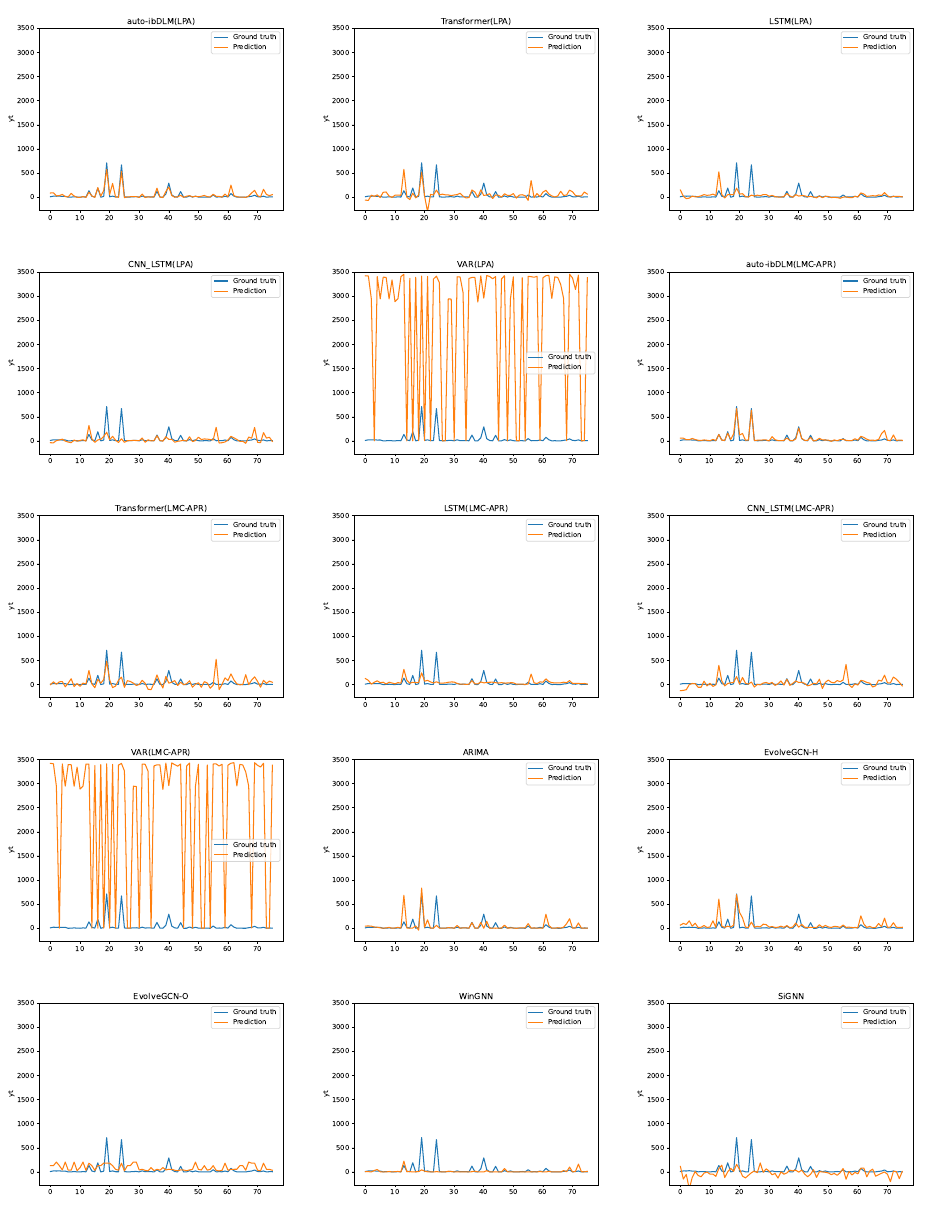}  
    \caption{Comparison of participant growth forecasting results on the Weibo benchmark. EvolveGCN, WinGNN, SiGNN, and ARIMA do not utilize community evolution behaviors and therefore produce identical results under LPA and LMC-APR. The horizontal axis represents the test sample index, and the vertical axis represents the predicted participant growth (i.e., node increment).}
\label{fig:comparat-model}
\end{figure}

Fig.~\ref{fig:comparat-model} compares the predicted participant growth (i.e., node increments) produced by different methods on the Weibo benchmark. The horizontal axis represents the test samples, each corresponding to a sliding window of nine consecutive network snapshots, while the vertical axis represents the participant growth to be predicted. Across both community detection schemes (LPA and LMC--APR), auto-ibDLM consistently produces predictions that are closest to the ground truth. In particular, it accurately captures major participant surges that are critical for timely public event forecasting, while overestimating only a few samples (e.g., samples 22, 62, 70, and 73 under LPA, and samples 22, 70, and 73 under LMC--APR).

Among the sequence models, Transformer captures some major growth trends but frequently overestimates participant growth (e.g., samples 14 and 57) and misses several obvious surges. LSTM and CNN--LSTM exhibit weaker sensitivity to sudden participant growth, often failing to detect true surges while occasionally generating false alarms. These results indicate that purely sequence-based models struggle to capture the highly nonlinear evolution of participant interactions.

The statistical models perform even less satisfactorily. VAR is unable to model the evolution of participant growth, whereas ARIMA captures several growth trends but tends to overestimate surge magnitudes and misses some prominent participant surges.

Among the dynamic graph representation learning methods, EvolveGCN-H captures the largest surge peaks but substantially underestimates participant growth in most cases. EvolveGCN-O fails to capture major surges, WinGNN consistently underestimates participant growth, and SiGNN cannot accurately model the overall evolution trend. These results suggest that directly learning graph representations from historical interaction networks is insufficient for accurately forecasting future participant growth.

Overall, auto-ibDLM achieves the most accurate participant growth forecasting performance on the Weibo benchmark. Its ability to simultaneously capture the overall evolution trend and sudden participant surges demonstrates the effectiveness of the proposed network-driven forecasting formulation and hybrid representation learning strategy.

\subsubsection{Representation-level interpretability analysis}
Unlike end-to-end graph representation learning methods, the proposed hybrid representation learning strategy provides representation-level interpretability by explicitly connecting network science-informed structural metrics with the learned latent representations. Since the auto-learning layer performs a linear projection from structural feature vectors to the latent representation space, its weight matrix naturally provides a quantitative measure of the contribution of each structural metric to the learned representations. In this analysis, only the network-level and node-level structural metrics are considered. Community evolution behaviors are excluded because their values depend on the adopted community detection algorithm.

\begin{figure}
    \centering  \includegraphics[width=\linewidth]{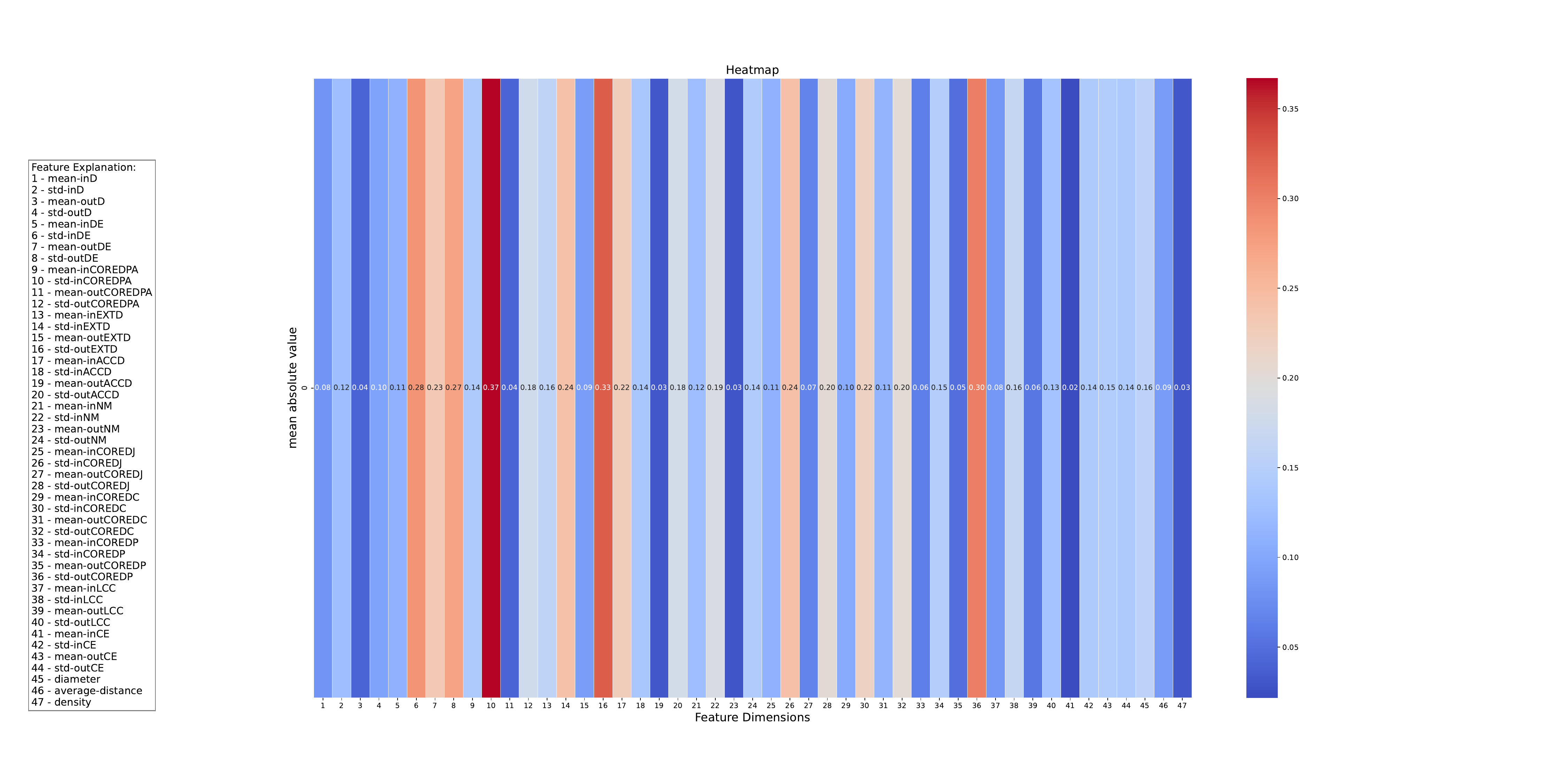} 
    \caption{Contribution scores of structural metrics derived from the auto-learning weight matrix (quantified by the mean absolute weight of each row).}
\label{fig:heatmap}
\end{figure}

To quantify the contribution of each structural metric, we compute the mean absolute value of the corresponding row in the auto-learning weight matrix, yielding a contribution score for every structural metric. Fig.~\ref{fig:heatmap} presents the resulting contribution heatmap. Higher scores indicate that the corresponding structural metrics contribute more substantially to the learned latent representations. The top three structural metrics are the standard deviations of in-COREDPA, out-EXTD, and out-COREDP, with contribution scores of 0.37, 0.33, and 0.30, respectively. Overall, twelve structural metrics achieve contribution scores greater than 0.2.

Interestingly, eleven of these twelve highly ranked structural metrics coincide with the manually identified informative structural metrics (listed in the Supplementary Information\footref{fn:supplementaryurl}), indicating that the learned latent representations largely preserve structural characteristics consistent with domain knowledge. The only exception is the standard deviation of in-COREDPA. Although its overall temporal trend appears weakly correlated with participant growth, it receives the highest contribution score, suggesting that the auto-learning layer successfully captures discriminative structural information that is difficult to identify through visual inspection alone. Conversely, some structural metrics exhibiting relatively consistent temporal trends (e.g., in-CE) receive comparatively low contribution scores, indicating that the learned latent representations capture more than simple temporal trend consistency.

To further validate the proposed contribution analysis, we remove the three highest-ranked structural metrics individually and jointly, retrain the model under the same experimental settings, and report the forecasting performance in Table~\ref{tab:contribution_validation}. Removing any of the three structural metrics consistently degrades the forecasting performance, while removing all three causes the largest deterioration, increasing the MSE from 0.000229 to 0.000528. Although the magnitude of performance degradation is not strictly consistent with the contribution ranking, this is expected because the proposed contribution scores quantify the average contribution of structural metrics to the learned latent representations rather than their isolated contributions to the final prediction. Moreover, redundancy and interactions among structural metrics allow part of the structural information to be compensated by the remaining metrics after removing an individual one. These results further demonstrate that the proposed contribution analysis successfully identifies structurally informative metrics that are important for learning effective latent representations.

\begin{table}
\centering
\caption{Validation of the contribution analysis by removing the top-ranked structural metrics, A lower MSE value indicates superior predictive performance.}
\label{tab:contribution_validation}
\begin{tabular*}{\textwidth}{@{\extracolsep{\fill}} lcc}
\toprule
Removed structural metric & MSE & Relative increase \\
\midrule
None & 0.000229 & -- \\
Std of in-COREDPA & 0.000311 & +35.8\% \\
Std of out-EXTD & 0.000425 & +85.6\% \\
Std of out-COREDP & 0.000387 & +69.0\% \\
Top-3 structural metrics & 0.000528 & +130.6\% \\
\botrule
\end{tabular*}
\end{table}

\begin{figure}
    \centering  \includegraphics[width=\linewidth]{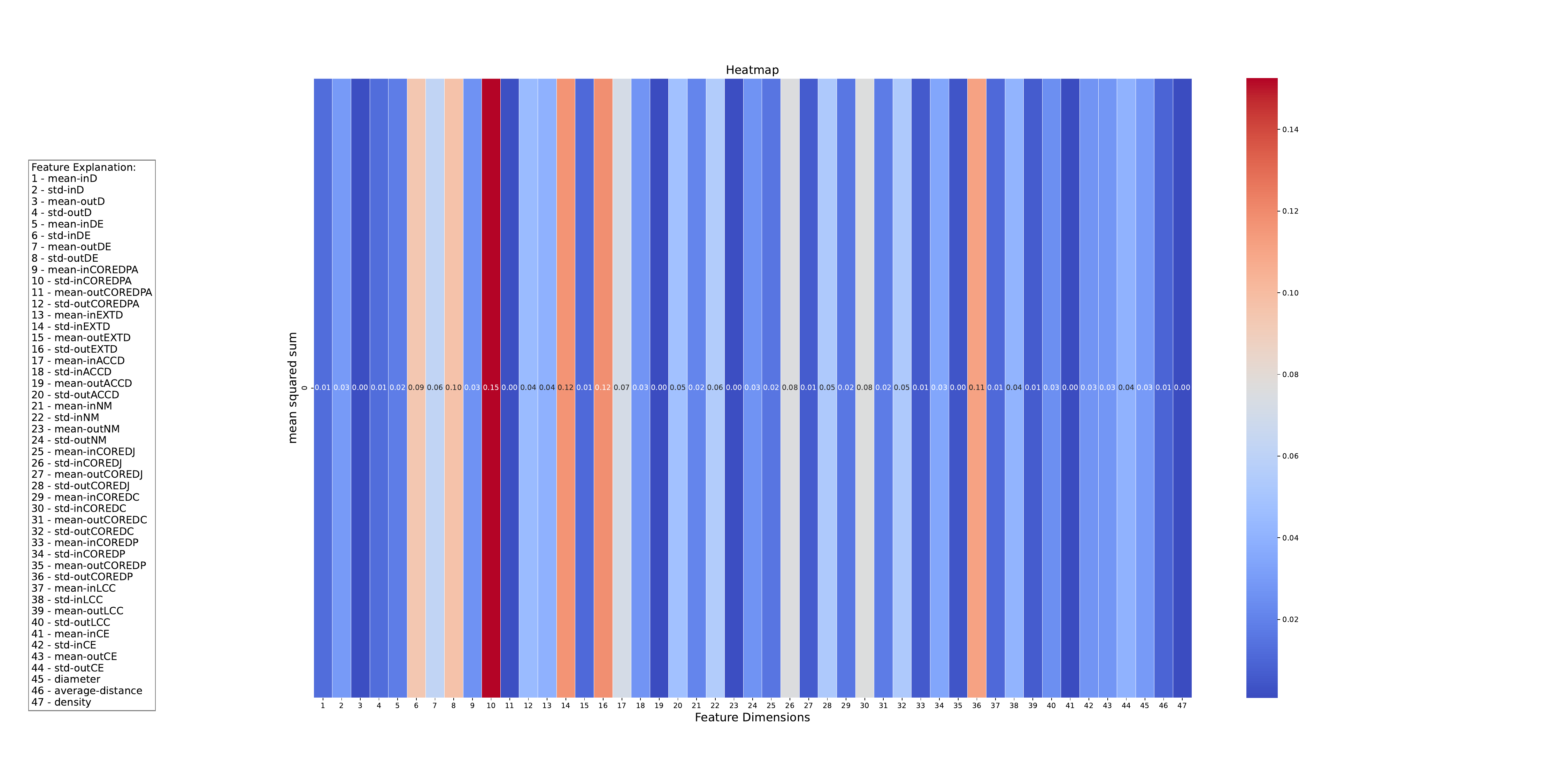} 
    \caption{Contribution scores of structural metrics derived from the auto-learning weight matrix (quantified by the mean squared sum per row).}
\label{fig:heatmap_sum}
\end{figure}

A complementary contribution analysis based on the mean squared weights is presented in Fig.~\ref{fig:heatmap_sum}. Although this criterion emphasizes structural metrics with high-magnitude contributions concentrated in a few latent dimensions, it produces a ranking highly consistent with that obtained using the mean absolute weights, further demonstrating the stability of the learned structural contributions.

Overall, the learned latent representations preserve structural information that is both consistent with established network science knowledge and capable of capturing subtle structural patterns beyond manual inspection. These results demonstrate that the proposed hybrid representation learning strategy provides meaningful representation-level interpretability while maintaining strong forecasting performance.

\subsubsection{Public event identification under different thresholds}
Although auto-ibDLM is trained to forecast participant growth rather than public event labels directly, public event forecasting can be naturally achieved by comparing the predicted participant growth against a predefined threshold. Specifically, an event is identified as a potential public event if its predicted participant growth exceeds the threshold. Table~\ref{tab:com-model-acc} reports the identification accuracies under thresholds of 200, 300, 400, 500, and 600 participants.

\begin{table}
    \caption{Accuracy (\%) of different models for public event identification under different participant growth thresholds.}
    \setlength{\tabcolsep}{2pt}
    \begin{tabular*}{\textwidth}{@{\extracolsep{\fill}} ccccccccccc}
        \toprule
        Baseline & \multicolumn{2}{c}{200}& \multicolumn{2}{c}{300}& \multicolumn{2}{c}{400}& \multicolumn{2}{c}{500}& \multicolumn{2}{c}{600}\\
    & \tiny LPA & \tiny LMC-APR & \tiny LPA & \tiny LMC-APR & \tiny LPA & \tiny LMC-APR & \tiny LPA & \tiny LMC-APR & \tiny LPA & \tiny LMC-APR\\ 
        \midrule
        auto-ibDLM & \textbf{97.3\%}  &\textbf{98.7\%} & \textbf{100\%}  &\textbf{100\%}  & \textbf{100\%}  &\textbf{100\%}  & \textbf{100\%}  & \textbf{100\%}  & \textbf{97.4\%} &\textbf{100\%}  \\
        Transformer & 94.7\% & 92.1\% & 96.1\% & 97.4\% & 97.4\% & 97.4\% & 97.4\% & 96.1\% & 97.4\% & 97.4\%
        \\
        LSTM & 94.7\%  &94.7\%  & 96.1\%  & 96.1\%  & 96.1\% & 97.4\%  & 96.1\%  & 97.4\%  & 97.4\%  & 97.4\% \\
        CNN-LSTM & 92.1\%  &  93.4\%  & 96.1\%  & 94.7\%  & 97.4\%  & 96.1\%  & 97.4\%  &97.4\%  & 97.4\%  &97.4\% \\ 
        VAR & 28.9\%  & 28.9\%  & 27.6\%  & 27.6\%  & 27.6\%  & 27.6\%  &  27.6\%  & 27.6\%  & 27.6\% & 27.6\% \\
        \midrule
        ARIMA & \multicolumn{2}{c}{94.7\%}& \multicolumn{2}{c}{97.4\%}& \multicolumn{2}{c}{97.4\%}& \multicolumn{2}{c}{97.4\%}& \multicolumn{2}{c}{97.4\%}\\
         EvolveGCN-H & \multicolumn{2}{c}{90.8\%}& \multicolumn{2}{c}{96.1\%}& \multicolumn{2}{c}{97.4\%}& \multicolumn{2}{c}{97.4\%}& \multicolumn{2}{c}{97.4\%}\\
        EvolveGCN-O & \multicolumn{2}{c}{85.5\%}& \multicolumn{2}{c}{97.4\%}& \multicolumn{2}{c}{97.4\%}& \multicolumn{2}{c}{97.4\%}& \multicolumn{2}{c}{97.4\%}\\
        WinGNN & \multicolumn{2}{c}{94.7\%}& \multicolumn{2}{c}{97.4\%}& \multicolumn{2}{c}{97.4\%}& \multicolumn{2}{c}{97.4\%}& \multicolumn{2}{c}{97.4\%}\\
        SiGNN & \multicolumn{2}{c}{96.1\%}& \multicolumn{2}{c}{97.4\%}& \multicolumn{2}{c}{97.4\%}& \multicolumn{2}{c}{97.4\%}& \multicolumn{2}{c}{97.4\%}\\
        \botrule
    \end{tabular*}
    \label{tab:com-model-acc}
\end{table}

Across both community detection schemes, auto-ibDLM consistently achieves the highest identification accuracy. Under LPA, it attains 97.3\% at a threshold of 200 participants, reaches 100\% accuracy at thresholds of 300, 400, and 500 participants, and decreases slightly to 97.4\% at a threshold of 600 participants. Under LMC-APR, it achieves 98.7\% at a threshold of 200 participants and maintains 100\% accuracy across all thresholds from 300 to 600 participants. The slight decrease in identification accuracy at the threshold of 600 participants under LPA is caused by the slight underestimation of the two largest participant surges, which results in both positive samples being predicted below the threshold.

Most baseline methods exhibit higher identification accuracy as the threshold increases because larger thresholds are less sensitive to moderate forecasting errors. Among the sequence models, LSTM and CNN-LSTM show a monotonic improvement with increasing thresholds, both reaching 97.4\% at a threshold of 600 participants but performing noticeably worse at lower thresholds. Transformer achieves relatively stable performance, with accuracies ranging from 92.1\% to 97.4\%. Among the statistical models, ARIMA reaches 97.4\% accuracy at thresholds of 300 participants and above but drops to 94.7\% at the lowest threshold, while VAR consistently performs poorly, with accuracies ranging from only 27.6\% to 28.9\%. The dynamic graph representation learning methods exhibit similar characteristics. EvolveGCN-H and EvolveGCN-O achieve relatively low accuracies at the lowest threshold (90.8\% and 85.5\%, respectively), but their performance gradually converges to 96.1\% and 97.4\% as the threshold increases. WinGNN and SiGNN show similar trends, achieving competitive accuracies at higher thresholds but noticeably lower performance at the threshold of 200 participants.

Since missing a potential public event is considerably more critical than issuing a false alarm in practical early-warning systems, recall provides a more informative measure of forecasting effectiveness. As reported in Table~\ref{tab:com-model-recall}, auto-ibDLM achieves nearly perfect recall across almost all thresholds, reaching 100\% under all thresholds except the highest threshold (600 participants) under LPA, where recall decreases to 0 because only two positive samples exist in the test set at this threshold, and both are predicted slightly below the threshold. In contrast, most competing methods exhibit substantially lower Recall. Transformer detects only a small proportion of public events, while LSTM, CNN-LSTM, EvolveGCN-O, WinGNN, and SiGNN fail to identify most public events across nearly all thresholds. ARIMA and EvolveGCN-H achieve moderate recall, successfully identifying only part of the public events. Although VAR also achieves perfect recall, this is attributed to the fact that it consistently produces substantially overestimated predictions of participant growth for almost all test samples (as shown in Fig.~\ref{fig:comparat-model}), resulting in extremely low precision. Therefore, its high recall does not translate into reliable public event forecasting. Detailed precision results are provided in the Supplementary Information\footref{fn:supplementaryurl}, where auto-ibDLM consistently achieves the best balance between recall and precision.

\begin{table}
    \caption{Recall (\%) of different models for public event identification under different participant growth thresholds.}
    \setlength{\tabcolsep}{2pt}
    \begin{tabular*}{\textwidth}{@{\extracolsep{\fill}} ccccccccccc}
    \toprule
        Baseline & \multicolumn{2}{c}{200}& \multicolumn{2}{c}{300}& \multicolumn{2}{c}{400}& \multicolumn{2}{c}{500}& \multicolumn{2}{c}{600}\\
    & \tiny LPA & \tiny LMC-APR & \tiny LPA & \tiny LMC-APR & \tiny LPA & \tiny LMC-APR & \tiny LPA & \tiny LMC-APR & \tiny LPA & \tiny LMC-APR\\ 
        \midrule
        auto-ibDLM & \textbf{100\%}  &\textbf{100\%} & \textbf{100\%}  &\textbf{100\%}  & \textbf{100\%}  &\textbf{100\%}  & \textbf{100\%}  & \textbf{100\%}  & 0 &\textbf{100\%}  \\
        Transformer & 33.3\% & 33.3\% & 50\% & 50\% & 50\% & 50\% & 50\% & 0 & 0 & 0
        \\
        LSTM & 0  & 33.3\%  & 0  & 0  & 0 & 0  & 0  & 0  & 0  & 0 \\
        CNN-LSTM & 0  &  0  & 0  & 0  & 0  & 0  & 0  & 0  & 0 & 0 \\ 
        VAR & \textbf{100\%} & \textbf{100\%}  & \textbf{100\%}  & \textbf{100\%}  & \textbf{100\%}  & \textbf{100\%}  &  \textbf{100\%}  & \textbf{100\%}  & \textbf{100\%} & \textbf{100\%} \\
        \midrule
        ARIMA & \multicolumn{2}{c}{33.3\%}& \multicolumn{2}{c}{50\%}& \multicolumn{2}{c}{50\%}& \multicolumn{2}{c}{50\%}& \multicolumn{2}{c}{50\%}\\
         EvolveGCN-H & \multicolumn{2}{c}{33.3\%}& \multicolumn{2}{c}{50\%}& \multicolumn{2}{c}{50\%}& \multicolumn{2}{c}{50\%}& \multicolumn{2}{c}{50\%}\\
        EvolveGCN-O & \multicolumn{2}{c}{0}& \multicolumn{2}{c}{0}& \multicolumn{2}{c}{0}& \multicolumn{2}{c}{0}& \multicolumn{2}{c}{0}\\
        WinGNN & \multicolumn{2}{c}{0}& \multicolumn{2}{c}{0}& \multicolumn{2}{c}{0}& \multicolumn{2}{c}{0}& \multicolumn{2}{c}{0}\\
        SiGNN & \multicolumn{2}{c}{0}& \multicolumn{2}{c}{0}& \multicolumn{2}{c}{0}& \multicolumn{2}{c}{0}& \multicolumn{2}{c}{0}\\
    \botrule
    \end{tabular*}
    \label{tab:com-model-recall}
\end{table}

Overall, auto-ibDLM consistently provides the most reliable public event identification performance across different thresholds. More importantly, its consistently high recall demonstrates a strong capability to identify potential public events before they evolve, validating the effectiveness of forecasting participant growth as the basis for public event forecasting.

\subsection{Node increment prediction on general dynamic networks}
\label{subsec:exp-p-nodeinc}
Table~\ref{tab:com-model} reports the node increment prediction performance on the Weibo, Superuser, and Bitcoin datasets in terms of MSE. For completeness, the Weibo results from the public event forecasting experiments are included, enabling a unified comparison across event-driven and general dynamic networks. Across all datasets and both community detection schemes, auto-ibDLM consistently achieves the lowest prediction errors. Specifically, it obtains MSEs of 0.000094 (LPA) and 0.000085 (LMC-APR) on Superuser, 0.000627 (LPA) and 0.000400 (LMC-APR) on Bitcoin, and 0.000126 (LPA) and 0.000071 (LMC-APR) on Weibo. These results demonstrate that the proposed framework generalizes well to dynamic networks exhibiting substantially different evolution characteristics.

\begin{table}
\caption{MSEs of different models across all datasets. The best results are highlighted in bold, and “-” indicates that the corresponding model fails to produce a valid prediction.}
\begin{tabular*}{\textwidth}{@{\extracolsep{\fill}} ccccccc}
\toprule
 Datasets & \multicolumn{2}{c}{Weibo} & \multicolumn{2}{c}{Superuser} & \multicolumn{2}{c}{Bitcoin} \\
 Models  & \tiny LPA & \tiny LMC-APR & \tiny LPA & \tiny LMC-APR & \tiny LPA & \tiny LMC-APR \\
 \midrule
 auto-ibDLM & \textbf{0.000126} & \textbf{0.000071} & \textbf{0.000094} & \textbf{0.000085} & \textbf{0.000627} & \textbf{0.000400} \\
 Transformer & 0.000521 & 0.000521 & 0.000154 & 0.000100 & 0.004592 & 0.003486
 \\
 LSTM & 0.000515 & 0.000442 & 0.000114 & 0.000111 & 0.003944 & 0.002000\\
 CNN-LSTM & 0.000547 & 0.000651 & 0.000259 & 0.000198 & 0.006087 & 0.006127\\
 VAR & 0.31161 & 0.31235 & 0.007623 & 0.007871 & 0.033059 & 0.038454\\
 ARIMA & \multicolumn{2}{c}{0.00050} & \multicolumn{2}{c}{0.00169} & \multicolumn{2}{c}{0.00925} \\
 EvolveGCN-H & \multicolumn{2}{c}{0.00049} & \multicolumn{2}{c}{0.00785} & 
 \multicolumn{2}{c}{0.00473} \\
 EvolveGCN-O & \multicolumn{2}{c}{0.00066} & \multicolumn{2}{c}{0.00789} & 
 \multicolumn{2}{c}{0.00814} \\
 WinGNN & \multicolumn{2}{c}{0.00056} & \multicolumn{2}{c}{0.00142} & \multicolumn{2}{c}{0.00568}\\
 SiGNN & \multicolumn{2}{c}{0.00078} & \multicolumn{2}{c}{0.00497} & \multicolumn{2}{c}{0.00789}\\
\botrule
\end{tabular*}
\label{tab:com-model}
\end{table}

Among the sequence models, LSTM and Transformer achieve relatively competitive performance on Superuser, obtaining MSEs of 0.000114 (0.000111) and 0.000154 (0.000100) under LPA (LMC-APR), respectively. However, their prediction errors increase considerably on Weibo and Bitcoin, where participant growth or node increment exhibits stronger fluctuations, indicating that purely sequence-based models struggle to model highly dynamic evolution patterns. CNN-LSTM consistently performs worse than both LSTM and Transformer across all datasets. Among the statistical models, ARIMA achieves acceptable performance on Superuser but deteriorates substantially on Bitcoin, while VAR consistently produces the largest prediction errors.

Among the dynamic graph representation learning methods, WinGNN consistently achieves lower prediction errors than EvolveGCN-O and SiGNN across all datasets, while EvolveGCN-H outperforms EvolveGCN-O on Weibo and Bitcoin and exhibits comparable performance on Superuser. Nevertheless, none of the dynamic graph representation learning methods matches the prediction accuracy of auto-ibDLM, indicating that directly learning graph representations from historical network snapshots is less effective than the proposed hybrid representation learning strategy for node increment forecasting.

The prediction curves provided in the Supplementary Information\footref{fn:supplementaryurl} further support these quantitative results. Consistent with the MSE comparisons, auto-ibDLM more accurately tracks the temporal evolution of node increments than the competing methods across both general dynamic network datasets.

Overall, these results demonstrate that auto-ibDLM consistently achieves superior forecasting performance across dynamic networks with substantially different evolution characteristics. This effectiveness and generalization capability arise from the proposed hybrid representation learning strategy, which combines network science-informed structural metrics with automatic representation learning to learn robust latent representations across event-driven and general dynamic networks.

\subsection{Ablation study}
\label{sebsec:exp-as}
We constructed six variants to evaluate the contribution of the proposed hybrid representation learning strategy and different categories of structural metrics: (1) removing the auto-learning layer (ibDLM); (2) removing $loss_f$ while retaining the auto-learning layer (auto-ibDLM$_{-loss_f}$); (3) replacing the auto-learning layer with manually selected structural metrics based on temporal trend consistency (manual-ibDLM), where the selected metrics are listed in the Supplementary Information\footref{fn:supplementaryurl}; (4) using only node-level and community evolution features (auto-ibDLM$_{c+l}$); (5) using only network- and node-level structural metrics (auto-ibDLM$_{m+l}$); and (6) using only node-level structural metrics (auto-ibDLM$_l$). Table~\ref{tab:ablation} summarizes the corresponding results.

\begin{table}
\caption{MSEs of auto-ibDLM and its variants across all datasets. The best results are highlighted in bold.}
\begin{tabular*}{\textwidth}{@{\extracolsep{\fill}} ccccccc}
\toprule
 Datasets & \multicolumn{2}{c}{Weibo} & \multicolumn{2}{c}{Superuser} & \multicolumn{2}{c}{Bitcoin} \\
 Models & \tiny LPA & \tiny LMC-APR & \tiny LPA & \tiny LMC-APR & \tiny LPA & \tiny LMC-APR \\
 \midrule
 auto-ibDLM & \textbf{0.000126} & \textbf{0.000071} & \textbf{0.000094} & \textbf{0.000084} & \textbf{0.000627} & \textbf{0.000400} \\
 ibDLM & 0.000325 & 0.000486 & 0.000107 & 0.000095 & 0.003973 & 0.001281 \\
 auto-ibDLM$_{-loss_f}$ & 0.000423 & 0.000373 & 0.000114 & 0.000101 & 0.001496 & 0.001105 \\
 manual-ibDLM & 0.000249 & 0.000225 & 0.000118 & 0.000095 & 0.002509 & 0.000764 \\
 auto-ibDLM$_{c+l}$ & 0.000287 & 0.000210 & 0.000094 & 0.000088 & 0.000652 & 0.000506\\
 auto-ibDLM$_{m+l}$ & \multicolumn{2}{c}{0.000229} & \multicolumn{2}{c}{0.000089} & \multicolumn{2}{c}{0.000656} \\
 auto-ibDLM$_{l}$ & \multicolumn{2}{c}{0.000227} & \multicolumn{2}{c}{0.000090} &  \multicolumn{2}{c}{0.000637} \\
\botrule
\end{tabular*}
\label{tab:ablation}
\end{table}

Across all datasets, auto-ibDLM consistently outperformed ibDLM, auto-ibDLM$_{-loss_f}$, and manual-ibDLM, demonstrating the effectiveness of the proposed hybrid representation learning strategy. Compared with ibDLM, the consistent performance improvement verifies the necessity of automatically transforming handcrafted structural feature vectors into compact latent representations. Compared with auto-ibDLM$_{-loss_f}$, the superior performance further confirms that the proposed consistency loss effectively guides the auto-learning layer to preserve informative temporal evolution patterns. Moreover, the superiority over manual-ibDLM indicates that automatically learning latent representations is more effective than manually selecting structural metrics solely based on temporal trend consistency. Although ibDLM and auto-ibDLM$_{-loss_f}$ outperform each other on different datasets, neither consistently matches the performance of auto-ibDLM, indicating that both the auto-learning layer and the proposed consistency loss are indispensable components of the proposed framework.

Among different structural metric configurations, auto-ibDLM$_{c+l}$ consistently outperformed auto-ibDLM$_l$ and auto-ibDLM$_{m+l}$ under LMC-APR across all datasets, whereas the three variants achieved comparable performance under LPA. This observation suggests that incorporating fine-grained community evolution behaviors can further improve forecasting performance when high-quality community detection results are available. Since LMC-APR identifies relatively small communities while LPA tends to merge many nodes into large communities, these results further indicate that the effectiveness of community evolution features depends on the quality of the underlying community detection algorithm. Nevertheless, the performance differences among auto-ibDLM$_l$, auto-ibDLM$_{m+l}$, and auto-ibDLM$_{c+l}$ remain relatively small.

Overall, the ablation results validate the effectiveness of the proposed hybrid representation learning strategy, confirming that both the auto-learning layer and the consistency loss contribute to the superior forecasting performance of auto-ibDLM. Moreover, auto-ibDLM$_l$ still achieves lower prediction errors than all competing methods reported in Table~\ref{tab:com-model}, demonstrating that node-level structural metrics alone are sufficient to achieve excellent forecasting performance. Since network-level structural metrics are computationally more expensive to compute and community evolution behaviors rely on external community detection algorithms, this finding provides a practical guideline for deploying auto-ibDLM in real-world public event forecasting systems.

\section{Conclusion}
This paper presented auto-ibDLM, a network-driven deep learning framework for public event forecasting. By representing an event as a dynamic interaction network and forecasting future participant growth, auto-ibDLM formulates public event forecasting as the prediction of dynamic interaction network evolution. The proposed framework employs a hybrid representation learning strategy that combines network science-informed structural metrics with automatic representation learning to learn compact, robust, and interpretable latent representations, which are subsequently modeled by a GRU-based temporal forecasting module to predict future participant growth.

Extensive experiments lead to three main findings. First, modeling public event forecasting as dynamic interaction network evolution forecasting provides an effective basis for forecasting public event evolution. auto-ibDLM consistently outperformed existing methods on 13 real-world Sina Weibo public event datasets and achieved reliable public event identification across different participant growth thresholds. Second, the proposed hybrid representation learning strategy effectively bridges network science-informed structural metrics and automatic representation learning, producing more informative latent representations than existing deep sequential and dynamic graph learning methods. The representation-level interpretability analysis further demonstrates that the learned latent representations are primarily driven by node-level structural metrics while automatically capturing informative structural patterns that are difficult to identify through manual feature engineering. Third, evaluations on general dynamic networks demonstrate that the proposed framework generalizes well across networks with substantially different evolution characteristics. Meanwhile, the ablation study reveals that although integrating multi-level structural metrics yields the best overall performance, node-level structural metrics alone are sufficient to achieve excellent forecasting performance, providing a practical and computationally efficient configuration for real-world deployment.

Despite these encouraging results, several limitations deserve further investigation. First, the current framework models participant interactions solely through their structural evolution without distinguishing interaction semantics (e.g., supportive versus opposing opinions). Although interaction intensity provides a strong indicator of event evolution, incorporating semantic information may further improve the estimation of actual participant behavior and consequently enhance public event forecasting accuracy. Second, although the proposed hybrid representation learning strategy effectively suppresses less informative structural metrics, the ablation study suggests that further improvements remain possible in exploiting complementary information among different categories of structural metrics. More adaptive representation learning mechanisms may therefore further improve the robustness and efficiency of the learned latent representations.

Future work will focus on addressing these limitations by integrating multi-modal information and advanced natural language processing techniques to jointly model interaction structure and interaction semantics. In addition, we will investigate more adaptive representation learning mechanisms capable of automatically identifying informative structural metrics while further suppressing redundant information. We hope this work provides a useful step toward intelligent public event forecasting systems that support proactive risk management, timely decision-making, and urban public safety.

\bibliography{sn-bibliography}

\end{document}